\pdfoutput=1

\documentclass[11pt]{article}

\usepackage[preprint]{coling}

\usepackage{times}
\usepackage{latexsym}
\usepackage{multirow}

\usepackage[T1]{fontenc}

\usepackage[utf8]{inputenc}

\usepackage{microtype}

\usepackage{inconsolata}

\usepackage{graphicx}

\usepackage{tabularx}

\title{Enhancing Persona Classification in Dialogue Systems: A Graph Neural Network Approach}

\author{Konstantin Zaitsev \\
  HSE / Moscow \\
  \texttt{kzaytsev@hse.ru} \\}

\begin{document}
\maketitle
\begin{abstract}
In recent years, Large Language Models (LLMs) gain considerable attention for their potential to enhance personalized experiences in virtual assistants and chatbots. A key area of interest is the integration of personas into LLMs to improve dialogue naturalness and user engagement. This study addresses the challenge of persona classification, a crucial component in dialogue understanding, by proposing a framework that combines text embeddings with Graph Neural Networks (GNNs) for effective persona classification. Given the absence of dedicated persona classification datasets, we create a manually annotated dataset to facilitate model training and evaluation. Our method involves extracting semantic features from persona statements using text embeddings and constructing a graph where nodes represent personas and edges capture their similarities. The GNN component uses this graph structure to propagate relevant information, thereby improving classification performance. Experimental results show that our approach, in particular the integration of GNNs, significantly improves classification performance, especially with limited data. Our contributions include the development of a persona classification framework and the creation of a dataset.
\end{abstract}

\section{Introduction}
Large Language Models (LLMs) gain significant traction in recent years, finding applications in various domains, particularly in personalized experiences such as virtual assistants and chatbots. In order to enhance the naturalness and engagement of these interactions, it is feasible to integrate personas into the LLMs. In our research, a persona is a personal fact about an individual, such as "I have 4 children" or "I play video games every day." Recent studies demonstrate that integrating personas into language models through prompts (\citealp{kwon-etal-2023-ground}; \citealp{xu-etal-2023-towards-zero}; \citealp{kasahara-etal-2022-building}), embeddings (\citealp{tang-etal-2023-enhancing-personalized}; \citealp{cheng-etal-2023-pal}), or even graph-based representations (\citealp{ju-etal-2022-learning}; \citealp{li-etal-2022-enhancing-knowledge}; \citealp{zhu-etal-2023-paed}) can substantially improve user engagement and make conversations more natural and personalized.

The classification of personas plays a crucial role in dialogue understanding and offers new avenues for improving conversation quality. Analyzing the distribution of persona labels across a dialogue can reveal patterns in what, when, and how personas are utilized in real conversations. This understanding is essential for replicating real dialogue patterns, thereby enhancing the personalized experience provided by chatbots.

However, persona classification presents significant challenges. To the best of our knowledge, there is no publicly available dataset specifically designed for our research. This lack of data necessitates the creation of a labelled dataset and the training of models on a limited amount of information. In light of these constraints, the use of graph-based representations appears to be a promising approach to improve the overall quality of persona classification.

Our proposed method utilizes text embeddings to extract semantic features from the persona statements. In addition to these features, we construct a graph representation of the personas, where nodes represent persona statements and edges capture similarities between them. The Graph Neural Network component then operates on this graph structure, allowing the model to utilize relevant information from the neighbouring nodes.

In this paper, we present the details of our proposed method, describe our approach to dataset creation and labeling, and evaluate the performance of our combined classifier against baseline methods.

Our contributions are as follows:
\begin{itemize}
    \item \textbf{A Framework for Persona Classification:} We develop a framework that combines the strengths of text embeddings and GNNs. This framework effectively captures the semantic of personal facts through embeddings while also leveraging the graph structure of persona nodes via GNNs. 
    \item \textbf{Creation of a Manually Annotated Persona Classification Dataset:} Given the lack of existing datasets related to persona classification, we contribute manually annotated dataset specifically designed for this task. This contribution not only facilitates the development of our proposed framework but also serves as a foundation for subsequent research and analysis of the utilisation of personas in dialogues.
\end{itemize}

\section{Related Work}
Graph Neural Networks (GNNs) are a class of deep learning models designed to process graphs. There are numerous GNN architectures, with notable examples including Graph Convolutional Network (GCN) \cite{DBLP:journals/corr/KipfW16}, Graph Attention Network (GAT) \cite{veličković2018graphattentionnetworks}, and GraphSAGE \cite{DBLP:journals/corr/HamiltonYL17}. These networks are demonstrated to perform well in tasks where the graph structure is a significant source of information. These tasks include the prediction of node (\citealp{9378063}; \citealp{huang2021combining}), graph (\citealp{9460814}; \citealp{DBLP:journals/corr/abs-2107-01410}), and link properties (\citealp{li2024kermitknowledgegraphcompletion}; \citealp{he2023mocosamomentumcontrastknowledge}; \citealp{wang-etal-2022-simkgc}). There is a number of benchmarks for evaluating the performance of GNNs (\citealp{DBLP:journals/corr/abs-2005-00687}; \citealp{huang2023temporalgraphbenchmarkmachine}).

In the context of text classification, it is important to derive meaningful features from textual data. Traditional approaches, including one-hot encodings, bag-of-words (BOW) and TF-IDF, are commonly employed in classification tasks (\citealp{lin-etal-2023-linear}; \citealp{8950616}; \citealp{das2018improvedtextsentimentclassification}; \citealp{galke-scherp-2022-bag}). Nevertheless, more sophisticated methodologies that employ neural networks demonstrate enhanced efficacy in recent years. In many cases, text embeddings are employed as feature extractors (\citealp{sun2020finetuneberttextclassification}; \citealp{2023}; \citealp{kementchedjhieva2023explorationencoderdecoderapproachesmultilabel}).

Text features, whether derived from TF-IDF or neural models, can be integrated as input features for GNNs. For example, a study \cite{yao2018graphconvolutionalnetworkstext} demonstrates strong performance in node classification tasks by constructing a heterogeneous graph where nodes represent words and articles, and edges are weighted by TF-IDF scores and Pointwise Mutual Information (PMI). However, the embeddings in this approach are not fine-tuned during training. Fine-tuning embeddings can be computationally expensive, which presents a challenge that the authors \cite{lin-etal-2021-bertgcn} address by training models in batches. They train a BERT encoder \cite{devlin-etal-2019-bert} and a GNN together for the node classification tasks, and this combination shows the best performance across approaches that fine-tune only text embeddings.

Personas can be classified based on their inherent characteristics. A paper \cite{gao-etal-2023-peacok} identifies five distinct persona classes, which were defined based on extensive analysis of existing literature. Another paper \cite{kumar-etal-2024-adding} proposes a similar categorization, further validating the consistency and relevance of these persona classes across different research efforts.

\section{Data}
\subsection{Dataset}
\begin{table*}
    \centering
    \begin{tabular}{llc}
        \hline
        \textbf{User} & \textbf{Utterance} & \textbf{Persona}  \\
        \hline
        bot\_0 & Hi. My name is mike. How are you? & My name is Mike. \\
        bot\_1 & Hey. I'm good. How are you? & \\
        bot\_0 & Not bad. I just got back from the pool I love swimming. & I love swimming. \\
        bot\_1 & Nice! I have a love for dogs, well all animals really. & I love dogs \\
        \hline
    \end{tabular}
    \caption{Dialogue Sample with Personas from MSC}
    \label{tab:msc}
\end{table*}

There are a limited number of datasets that contain dialogues with annotated personas. Examples of such datasets include PersonaChat \cite{zhang-etal-2018-personalizing} and Multi-Session Chat (MSC) \cite{xu-etal-2022-beyond}. Both datasets consist of conversations between users, where each dialogue participant is associated with a set of personas. In contrast to PersonaChat, MSC is divided into multiple sessions, simulating real-life scenarios where conversations may be interrupted by time intervals. The MSC dataset is the most suitable for our purposes because it contains fairly simple annotation in which a user's utterance can contain a persona. Table \ref{tab:msc} presents a dialogue sample from the MSC dataset.

Although personas can be represented in text form, if the classifier only works with text embeddings, such a representation may lack sufficient informativeness. The paper \cite{gao-etal-2023-peacok} proposes the use of personas in a knowledge graph representation (PeaCoK). In this graph, personas are nodes and edges represent relations between them. Furthermore, they propose a classification of these relations based on works devoted to the analysis of human conversation.

Nevertheless, it is also feasible to categorize the nodes themselves, rather than merely the relations between them. In our approach, the persona graph is presented in a slightly simplified manner compared to how it would be constructed as a knowledge graph. It thus follows that our graph is undirected. This structure allows the creation of edges between personas in such a way that the most semantically close personas will be neighbours. Since semantically close personas are more likely to have the same classes, such a structure will allow the problem of classifying personas to be solved more effectively. We follow to the persona classification described in PeaCoK:

\begin{itemize}
    \item \textbf{Characteristics} describe mental state, personal traits, preferences, moods.
    \item \textbf{Experiences} describe events that occurred once or in the past. It may relate to the current work, marital status, etc.
    \item \textbf{Routines or Habits} describe frequently performed activities or activities the person does on a regular basis.
    \item \textbf{Goals or Plans} describe a person's wishes or actions that they want to achieve in the future.
    \item \textbf{Relationship} describes interactions with the other people. This label may overlap with other labels in cases where the utterance contains a persona about another person. For example, the persona "My sister wants to go to university" has the labels "Relationship" and "Goals or plans".
\end{itemize}

As can be seen with the "Relationship" label, each persona can have multiple labels. For example, a single persona can be related to both the "Experiences" and "Characteristics" labels, as shown in the examples in the Table \ref{tab:labels}.

\begin{table*}
    \centering
    \begin{tabular}{lc}
    \hline
    \textbf{Persona} & \textbf{Labels} \\
    \hline
    A lot of my family members are teachers. & Relationship, Experiences \\
    I am afraid of snakes. & Characteristics \\
    i live alone with my cats. & Experiences \\
    Each morning I make an omelet with 6 eggs. & Routines or Habits \\
    I am married with two children. & Characteristics, Experience \\
    I want to buy a muscle car. & Goals or Plans \\
    \hline
    \end{tabular}
    \caption{Labeled Persona Samples}
    \label{tab:labels}
\end{table*}

\subsection{Persona Labeling}
Given that PeaCoK employs the use of classes to represent the relationships between personas, whereas our study necessitates the classification of nodes, we devise our own dataset. In this dataset we annotate labels that we define in the previous paragraph. However, the manual annotation process can be expensive. There are studies (\citealp{tan2024largelanguagemodelsdata}; \citealp{rouzegar-makrehchi-2024-enhancing}; \citealp{aguda-etal-2024-large}) that propose to collect annotations using LLMs. The studies demonstrate that this approach can assist annotators in the collection of high-quality annotations.

According to this fact, we use LLM\footnote{\url{https://huggingface.co/macadeliccc/WestLake-7B-v2-laser-truthy-dpo}} for the annotation process. We develop a detailed prompt that describes the task and provides label descriptions. To improve the quality of the classification, we add several examples to the task description, covering all classes. We provide a list of personas as input. As a result, we expect a response in JSON format, where the keys are classes and the values are a list of personas. Thus, we get the structured response. The complete prompt is available in Appendix ~\ref{sec:prompt}.

\begin{table}
    \centering
    \begin{tabular}{lcc}
    \hline
    \textbf{Label} & \textbf{Train} & \textbf{Test} \\
    \hline 
    Experiences & 1368 & 263 \\
    Characteristics & 977 & 218 \\
    Routines or Habits & 272 & 87 \\
    Goals or Plans & 112 & 76 \\
    Relationship & 160 & 32 \\
    \hline
    \textbf{Overall} & 2889 & 676 \\
    \hline
    \end{tabular}
    \caption{Dataset Statistics}
    \label{tab:statistics}
\end{table}

After writing the prompt, we consistently collect annotations using LLM on all data from the MSC dataset. Next, we manually validate the collected annotations because LLM can make mistakes. The validation is done using Label Studio\footnote{\url{https://labelstud.io}}. During the validation process we find that LLM is wrong in 1 out of 5 cases. 

To reduce the number of errors and speed up the validation process, we iteratively train the classifier on the collected personas. The iteration is similar to active learning. We validate 100 personas, add validated personas to the already validated dataset, train the classifier, predict classes for unvalidated personas and validate the predictions again.

We manually validate and annotate around 3,000 personas. This amount is sufficient for our experiments. This is due to the fact that the MSC dataset very often repeats the meaning of the same personas. As a result of the annotation process, we get a dataset in which each person has multiple labels, as shown in the Table \ref{tab:labels}. The final statistics of the dataset are shown in the Table \ref{tab:statistics}. 

\section{Approach}
\subsection{Graph Creation}
TextGCN and BertGCN, which are based on text features and GNN, utilise heterogeneous graphs that connect words to articles. In contrast, our approach focuses on a homogeneous graph where edges are created between personas. Additionally, the existing studies propose weighting according to PMI or TF-IDF scores. However, in the context of persona classification, such an approach may not be as relevant. 

The issue is that, rather than focusing on the statistics, it is necessary to consider the semantic similarity between the personas. This is because, in some cases, personas with opposing meanings may have high TF-IDF scores but be semantically distant. For instance, the phrases "I like ice cream" and "I don't like ice cream" would have high TF-IDF scores but low scores for the semantic model if the approach and model are selected appropriately.

A second issue is that the process of computing similarities between nodes can be time-consuming due to the fully connected nature of the graph. However, there is no need to count all pairs, as it is evident that some personas will undoubtedly exhibit significant semantic differences. To address these challenges, we propose a graph construction process.

\textbf{Embeddings Extraction}. Firstly, we obtain text embeddings for personas. We select the e5-large text embedding model\footnote{\url{https://huggingface.co/intfloat/multilingual-e5-large}} as it achieves a high ranking on the MTEB\footnote{\url{https://huggingface.co/spaces/mteb/leaderboard}} leaderboard.

\textbf{Neighbor Identification}. Subsequently, we employ the k-nearest neighbours (k-NN) algorithm to ascertain the nearest neighbours for each persona. We select cosine similarity as the metric. It is noteworthy that the number of neighbours is a hyperparameter that can be adjusted. In our experiments, the optimal choice is seven neighbours.

\textbf{Edge Weighting}. 
Within each group of neighbours, we compute pairwise similarity scores using Natural Language Inference (NLI). We choose NLI because similarity scores do not always take into account opposite personas. In this formulation, we construct all neighbouring pairs of personas and compute the entailment score using the DeBERTa classifier\footnote{\url{https://huggingface.co/MoritzLaurer/DeBERTa-v3-base-mnli-fever-anli}}. We then apply these scores as edge weights.

\textbf{Graph Construction}. Finally, we construct a weighted graph where the nodes represent personas and the edges are weighted by the similarity scores obtained from the NLI model. We retain all edges regardless of weight, as we hypothesise that even small edge weights provide valuable information for classification.

\begin{figure}[t]
  \includegraphics[width=\columnwidth]{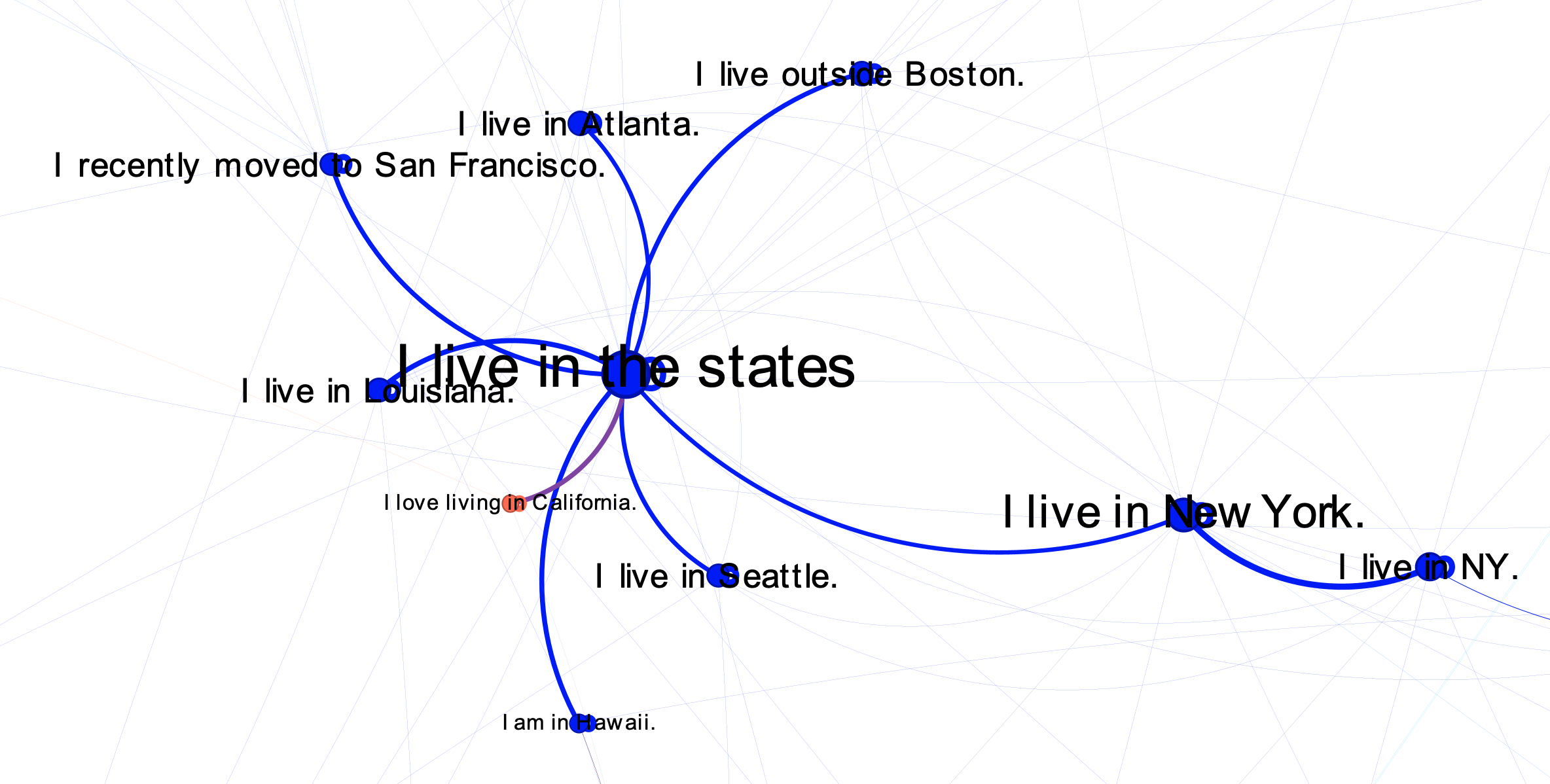}
  \caption{An example of a subgraph. Here the blue color is the label "Experiences", the red color is the multilabel "Experiences" and "Characteristics".}
  \label{fig:subgraph}
\end{figure}

The resulting graph is a weighted homogeneous one. The proposed approach allows for the construction of graphs in a relatively short time while ensuring the accurate preservation of semantic similarities between personas. Figure \ref{fig:subgraph} illustrates a sample of the constructed graph.

\subsection{Model Architectures}
We conduct experiments with different approaches and architectures. The aim is to show that graph structure helps simple models to improve their quality even having small amount data samples. To prove this hypothesis, we train small models with and without additional layers using GNN. We want to show the influence of the graph representation on the quality metrics. Since a persona can belong to several labels, the task for the models is to solve a multi-label classification.

\subsubsection{Linear Baselines}
As a baseline, we use features produced by TF-IDF, BOW and pre-trained text embeddings. We use these features as input to a classifier based on logistic regression and do not fine-tune the embeddings in this case. This baseline serves as a simple point of comparison for more complex models.

\subsubsection{Text Features with GNN Training}
In this setup, we use the same features as in the linear baselines, but replace the logistic regression with a GNN. The final layer of the GNN is a linear projection that outputs label probabilities. Importantly, we do not fine-tune the embeddings in this model either. We train only the GNN. This allows us to evaluate the effect of training on the embedding layer.

\subsubsection{Text Embeddings Fine-Tuning}
The next step is to fine-tune the text embeddings for the classification task. We use a text embedding model and a final linear layer that predicts probabilities of persona labels. This model helps us to understand the importance of including a GNN in addition to the fine-tuned embeddings.

\subsubsection{Text Embeddings Fine-Tuning with GNN Training}
In the final model, we fine-tune the text embeddings and train the GNN together. We hypothesise that training both models can improve the quality of the model. This is because the text embeddings are tuned for the task and GNN provides additional information as the personas with the same labels are likely to be neighbours. 

Often GNN needs to be trained on all data samples. Although this is not a problem for GNN, fine-tuning the text embeddings on all samples is computationally expensive. We follow BertGCN and add batches in the training process. At each epoch we obtain text embeddings, store them in a matrix, and update the matrix during the forward pass. It is important to note that we do not fine-tune the matrix, but fine-tune the text embedding model. 

The model is designed to classify text by leveraging both a pre-trained text encoder and a GNN. The encoder captures the semantic content of the text, while the GNN incorporates graph-based relational information. The output logits from both components are combined using a weighted sum to produce the final classification logits.

The encoder component uses a pre-trained model (e.g., BERT) to extract semantic features from the input text. The feature extraction can be represented as:

\begin{equation}
    H = Encoder(X)
\end{equation}

In this equation, $H$ is the hidden representation obtained from the encoder for the input tokens $X$. The extracted features $H$ are fed into a classifier head to produce the classification logits ($Z_{Encoder}$):

\begin{equation}
    Z_{Encoder} = ClassifierHead(H)
\end{equation}

The $ClassifierHead$ consists of a dense layer followed by a dropout layer and an output projection layer.

The $GNN$ component takes the encoder features and additional graph-based information to compute another set of classification logits ($Z_{GNN}$). The $GNN$ utilizes node features, edge index, and edge weight:

\begin{equation}
    Z_{GNN} = GNN(H, edge_{index}, edge_{weight})
\end{equation}

The final logits ($Z$) are obtained by combining ($Z_{Encoder}$) and ($Z_{GNN}$) using a weighted sum, controlled by the hyperparameter ($\lambda$):

\begin{equation}
    Z = \lambda \cdot Z_{GNN} + (1 - \lambda) \cdot Z_{Encoder}
\end{equation}

We set ($\lambda$) to 0.7, following the implementation inspired by BertGCN.

\begin{table*}
    \centering
    \begin{tabular}{cl|cc|cc|cc}
        \hline
        \textbf{\% train} & \textbf{model} & \textbf{F1} & \textbf{std} & \textbf{Precision} & \textbf{std} & \textbf{Recall} & \textbf{std} \\
        \hline
        \multirow{6}{*}{1\%} & Pre-bge \& GraphSAGE & 0.5054 & 0.0086 & 0.5512 & 0.0400 & 0.6231 & 0.0178  \\
        & TF-IDF \& GraphSAGE & 0.4855 & 0.0050 & 0.3577 & 0.0079 & 0.8038 & 0.0028 \\
        & FT-bge \& GraphSAGE & 0.4770 & 0.0100 & 0.3658 & 0.0130 & 0.7249 & 0.0813  \\
        & FT-bge \& Linear & 0.4740 & 0.0000 & 0.3253 & 0.0000 & 1.0000 & 0.0000 \\
        & Pre-bge \& Linear & 0.4582 & 0.1610 & 0.4268 & 0.1500 & 0.5445 & 0.1913  \\
        & TF-IDF \& Linear & 0.4202 & 0.1477 & 0.3182 & 0.1118 & 0.6297 & 0.2213 \\
        \hline
        \multirow{6}{*}{30\%} & FT-bge \& GraphSAGE & 0.8325 & 0.0134 & 0.8085 & 0.0154 & 0.8790 & 0.0072 \\
        & Pre-bge \& Linear & 0.7743 & 0.0077 & 0.7554 & 0.0136 & 0.8189 & 0.0019 \\
        & Pre-bge \& GraphSAGE & 0.7663 & 0.0082 & 0.7504 & 0.0087 & 0.8115 & 0.0028 \\
        & FT-bge \& Linear & 0.7127 & 0.0213 & 0.6560 & 0.0218 & 0.7985 & 0.0150 \\
        & TF-IDF \& GraphSAGE & 0.6049 & 0.0173 & 0.6285 & 0.0162 & 0.7166 & 0.0065 \\
        & TF-IDF \& Linear & 0.5613 & 0.0017 & 0.6225 & 0.0349 & 0.6331 & 0.0047 \\
        \hline
        \multirow{6}{*}{100\%} & FT-bge \& GraphSAGE & 0.8872 & 0.0007 & 0.8668 & 0.0013 & 0.9093 & 0.0024 \\
        & FT-bge \& Linear & 0.8742 & 0.0015 & 0.8507 & 0.0018 & 0.9012 & 0.0009 \\
        & Pre-bge \* GraphSAGE & 0.8258 & 0.0017 & 0.7929 & 0.0007 & 0.8697 & 0.0051 \\
        & Pre-bge \& Linear & 0.8234 & 0.0027 & 0.8058 & 0.0045 & 0.8515 & 0.0019 \\
        & TF-IDF \& GraphSAGE & 0.6821 & 0.0038 & 0.6334 & 0.0224 & 0.7868 & 0.0239 \\
        & TF-IDF \& Linear & 0.6729 & 0.0067 & 0.6948 & 0.0073 & 0.7306 & 0.0005 \\
        \hline
    \end{tabular}
    \caption{Persona Classification Metrics}
    \label{tab:metrics}
\end{table*}

\subsubsection{Experiments Setup}
We use \textit{GraphSAGE} as our GNN because of its superior performance compared to other GNN architectures. The sequence length for the text embedding model is set to 256 tokens, with a batch size of 32. The learning rates are set differently for each component to optimise convergence: 2e-4 for the encoder and 2e-3 for the GNN. We use the \textit{bge-small}\footnote{\url{https://huggingface.co/BAAI/bge-small-en-v1.5}} model as the text embedding model, which is lightweight enough for fast hypothesis testing.

To study the effect of adding GNNs in detail, we split the dataset into different ratios and perform 10 runs for each split. All models are trained for 20 epochs, except for models where we do not fine-tune the embeddings, which are trained for 1000 epochs to account for their slower convergence. To avoid overfitting, we select the best model based on the highest F1 score.

In order to evaluate the performance of the models, we select three metrics: F1, precision and recall. These metrics are chosen due to the fact that the dataset presents an imbalanced distribution of labels. For each model, we identify the optimal threshold for the probabilities, which corresponds to the highest F1 score. Given that the task is that of multi-label classification task, we choose Binary Cross-Entropy with Logits as the loss function. 

\section{Results}
Table \ref{tab:metrics} provides a detailed comparison of persona classification metrics across different models and training dataset sizes. The column labelled "\% train" indicates the percentage of the total training samples used for model training. The prefixes "Pre" and "FT" indicate the use of pre-trained and fine-tuned \textit{bge-small} models respectively. Note that TF-IDF and BOW are only used as input features in all models, whether for GNN or linear classifiers, and are not subject to fine tuning. The complete metrics can be found in the appendix \ref{sec:metrics}.

The analysis reveals that the integration of the \textit{bge-small} model with the \textit{GraphSAGE} architecture consistently yields superior performance across varying dataset sizes. This combination demonstrates remarkable efficacy even with limited data, achieving high-quality results at a mere 30\% dataset size. In contrast, the standalone fine-tuned \textit{bge-small} model requires the full dataset (100\%) to reach comparable performance levels.

At full dataset training (100\%), both the \textit{FT-bge \& GraphSAGE} and the fine-tuned \textit{bge-small} models achieve almost identical performance, with F1 scores of 0.88 and 0.87 respectively. This suggests that while \textit{GraphSAGE} has a significant advantage in the early stages of training, its impact diminishes as the dataset size increases.

The results emphasize that the graph-based approach (\textit{GraphSAGE}) promotes faster convergence and increases model quality, especially when data is sparse. However, as the dataset becomes more comprehensive, the advantages of incorporating a graph structure become less pronounced, allowing simpler models to achieve comparable performance.

In conclusion, these results underline the value of graph-based methods in scenarios characterised by data scarcity, while also highlighting that simpler models can perform equally well when data is abundant. The proposed method, combining \textit{bge-small} with \textit{GraphSAGE}, provides a solution for efficient classification in different data availability scenarios.

\section{Limitations}
While our proposed persona classification framework shows promising results, several limitations must be acknowledged.

First, despite the use of LLM for initial labelling, the annotation process still requires significant human intervention for validation and correction. Manual validation revealed that LLMs made errors in about 20\% of cases, highlighting potential weaknesses in automated annotation processes. In addition, the inherent subjectivity of persona classification can lead to inconsistencies in labelling, even after human validation. This subjectivity can result in noisy labels, which can affect the performance of classification models.

Second, although carefully constructed, our dataset is relatively small and may not capture the full diversity and complexity of personas as they occur in real-world dialogues. Our reliance on a single dataset, the MSC dataset, limits the generalisability of our findings. Despite our efforts to ensure a diverse representation of personas, the dataset may still lack coverage of certain persona types or interactions found in broader conversational contexts.

Finally, while the graph-based approach is effective in improving classification performance with limited data, it introduces additional computational complexity. The construction of the graph and the calculation of semantic similarities between persona statements can be resource intensive, especially as the size of the dataset increases. This may limit the scalability of our approach when applied to larger datasets or in real-time applications.

In conclusion, while our framework advances the field of persona classification, future work should focus on addressing these limitations by exploring more robust annotation techniques, expanding the diversity of datasets and improving computational efficiency.

\bibliography{custom}


\appendix

\clearpage
\newpage
\section{Prompt Example}
\label{sec:prompt}

\begin{table}[h!]
    \centering
    \begin{tabularx}{\textwidth}{X}
        \hline
        \textbf{Prompt} \\
        \hline
         You have a persona list of a person participating in dialogue. You should classify the following list into categories:
         
        Characteristics describe an intrinsic trait, e.g., a quality or a mental state, that the persona likely exhibits.

        Routines or Habits describe an extrinsic behaviour that the persona does on a regular basis.
        
        Goals or Plans describe an extrinsic action or outcome that the persona wants to accomplish or do in the future.
        
        Experiences describe extrinsic events or activities that the persona did in the past or does in the present.
        
        Do not add any commentaries. Return the answer in JSON format. All statements should be classified.
        
        Example:
        
        "I like to hike.", "I don't have kids.", "I have a neighbor named John.", "I don't have a job right now.", "I quit my previous job.", "I don't like to drink."
        
        Output:
        
        \{
        "Characteristics": ["I don't have kids.", "I don't like to drink."],
        "Routines or Habits": ["I like to hike.", ],
        "Goals or Plans ": [],
        "Experiences": ["I quit my previous job.", "I have a neighbor named John.", "I don't have a job right now."]
        \}
        
        Example:
        
        "I have to use my money carefully.", "I like meat.", "Chicken is my favorite meat.", "I like painting and composing music.", "I'm not a big fan of science fiction."
        
        Output:
        
        \{
        "Characteristics": ["I have to use my money carefully.", "I like meat.", "Chicken is my favorite meat.", "I'm not a big fan of science fiction."],
        "Routines or Habits": ["I like painting and composing music."],
        "Goals or Plans ": [],
        "Experiences": []
        \}
        
        Example:
        "I like to work on cars.", "Fords are my favourite cars.", "I plan to go back and finish college.", "I grew up in a small town.", "I have pets.", "I have cats and dogs.", "I work in a grocery store.", "I want a better job."
        
        Output:
        
        \{
        "Characteristics": ["Fords are my favourite cars."],
        "Routines or Habits": ["I like to work on cars."],
        "Goals or Plans ": ["I plan to go back and finish college.", "I want a better job."],
        "Experiences": ["I grew up in a small town.", "I have pets.", "I have cats and dogs.", "I work in a grocery store."]
        \}
        
        Example:
        
        "I had a burger and chocolate milkshake for my lunch.", "I work for jnj.", "I am going snowboarding.", "I am buying a new car."
        
        Output:
        
        \{
        "Characteristics": [],
        "Routines or Habits": ["I am going snowboarding."],
        "Goals or Plans ": ["I am buying a new car."],
        "Experiences": ["I had a burger and chocolate milkshake for my lunch.", "I work for jnj."]
        \}
        
        Example:
        
        \{samples\}
        
        Output: \\
        \hline
    \end{tabularx}
    \caption{Prompt Example}
    \label{tab:prompt}
\end{table}

\clearpage
\section{Full Metrics}
\label{sec:metrics}

\begin{table}[h!]
    \centering
    \begin{tabular}{cl|cc|cc|cc}
        \hline
        \textbf{\% train} & \textbf{model} & \textbf{F1} & \textbf{std} & \textbf{Precision} & \textbf{std} & \textbf{Recall} & \textbf{std} \\
        \hline
        \multirow{8}{*}{1\%} & Pre-bge \& GraphSAGE & 0.5054 & 0.0086 & 0.5512 & 0.0400 & 0.6231 & 0.0178  \\
        & TF-IDF \& GraphSAGE & 0.4855 & 0.0050 & 0.3577 & 0.0079 & 0.8038 & 0.0028 \\
        & BOW \& GraphSAGE & 0.4830 & 0.0046 & 0.4229 & 0.0355 & 0.6583 & 0.1076 \\
        & FT-bge \& GraphSAGE & 0.4770 & 0.0100 & 0.3658 & 0.0130 & 0.7249 & 0.0813  \\
        & FT-bge & 0.4740 & 0.0000 & 0.3253 & 0.0000 & 1.0000 & 0.0000 \\
        & Pre-bge \& Linear & 0.4582 & 0.1610 & 0.4268 & 0.1500 & 0.5445 & 0.1913  \\
        & BOW \& Linear & 0.4246 & 0.1492 & 0.3040 & 0.1068 & 0.7243 & 0.2545  \\
        & TF-IDF \& Linear & 0.4202 & 0.1477 & 0.3182 & 0.1118 & 0.6297 & 0.2213 \\
        \hline
        \multirow{8}{*}{30\%} & FT-bge \& GraphSAGE & 0.8325 & 0.0134 & 0.8085 & 0.0154 & 0.8790 & 0.0072 \\
        & Pre-bge \& Linear & 0.7743 & 0.0077 & 0.7554 & 0.0136 & 0.8189 & 0.0019 \\
        & Pre-bge \& GraphSAGE & 0.7663 & 0.0082 & 0.7504 & 0.0087 & 0.8115 & 0.0028 \\
        & FT-bge & 0.7127 & 0.0213 & 0.6560 & 0.0218 & 0.7985 & 0.0150 \\
        & BOW \& GraphSAGE & 0.6470 & 0.0132 & 0.6883 & 0.0052 & 0.7164 & 0.0061 \\
        & BOW \& Linear & 0.6415 & 0.0040 & 0.6284 & 0.0171 & 0.7602 & 0.0285 \\
        & TF-IDF \& GraphSAGE & 0.6049 & 0.0173 & 0.6285 & 0.0162 & 0.7166 & 0.0065 \\
        & TF-IDF \& Linear & 0.5613 & 0.0017 & 0.6225 & 0.0349 & 0.6331 & 0.0047 \\
        \hline
        \multirow{8}{*}{50\%} & FT-bge \& GraphSAGE & 0.8826 & 0.0023 & 0.8584 & 0.0053 & 0.9092 & 0.0056 \\
        & FT-bge & 0.8330 & 0.0048 & 0.8332 & 0.0105 & 0.8494 & 0.0009 \\
        & Pre-bge \& Linear & 0.8153 & 0.0044 & 0.7627 & 0.0085 & 0.8820 & 0.0178 \\
        & Pre-bge \& GraphSAGE & 0.8056 & 0.0001 & 0.7777 & 0.0027 & 0.8442 & 0.0014 \\
        & BOW \& GraphSAGE & 0.6734 & 0.0022 & 0.6970 & 0.0094 & 0.7204 & 0.0047 \\
        & BOW \& Linear & 0.6664 & 0.0014 & 0.6934 & 0.0021 & 0.7132 & 0.0005 \\
        & TF-IDF \& GraphSAGE & 0.6065 & 0.0069 & 0.6797 & 0.0105 & 0.6568 & 0.0140 \\
        & TF-IDF \& Linear & 0.5881 & 0.0001 & 0.6761 & 0.0075 & 0.6541 & 0.0009 \\
        \hline
        \multirow{8}{*}{70\%} & FT-bge \& GraphSAGE & 0.8819 & 0.0007 & 0.8661 & 0.0041 & 0.9004 & 0.0053 \\
        & FT-bge & 0.8497 & 0.0005 & 0.8280 & 0.0051 & 0.8772 & 0.0047 \\
        & Pre-bge \& GraphSAGE & 0.8209 & 0.0061 & 0.7929 & 0.0067 & 0.8608 & 0.0051 \\
        & Pre-bge \& Linear & 0.8089 & 0.0034 & 0.7783 & 0.0050 & 0.8487 & 0.0014 \\
        & BOW \& Linear & 0.7032 & 0.0009 & 0.6754 & 0.0080 & 0.7784 & 0.0103 \\
        & BOW \& GraphSAGE & 0.6866 & 0.0037 & 0.7539 & 0.0156 & 0.7049 & 0.0117 \\
        & TF-IDF \& GraphSAGE & 0.6518 & 0.0068 & 0.6633 & 0.0141 & 0.7396 & 0.0187 \\
        & TF-IDF \& Linear & 0.5979 & 0.0085 & 0.7786 & 0.0492 & 0.6519 & 0.0220 \\
        \hline
        \multirow{8}{*}{100\%} & FT-bge \& GraphSAGE & 0.8872 & 0.0007 & 0.8668 & 0.0013 & 0.9093 & 0.0024 \\
        & FT-bge & 0.8742 & 0.0015 & 0.8507 & 0.0018 & 0.9012 & 0.0009 \\
        & Pre-bge \* GraphSAGE & 0.8258 & 0.0017 & 0.7929 & 0.0007 & 0.8697 & 0.0051 \\
        & Pre-bge \& Linear & 0.8234 & 0.0027 & 0.8058 & 0.0045 & 0.8515 & 0.0019 \\
        & BOW \& Linear & 0.7211 & 0.0060 & 0.7306 & 0.0054 & 0.7561 & 0.0042 \\
        & BOW \& GraphSAGE & 0.7129 & 0.0026 & 0.7293 & 0.0049 & 0.7324 & 0.0098 \\
        & TF-IDF \& GraphSAGE & 0.6821 & 0.0038 & 0.6334 & 0.0224 & 0.7868 & 0.0239 \\
        & TF-IDF \& Linear & 0.6729 & 0.0067 & 0.6948 & 0.0073 & 0.7306 & 0.0005 \\
        \hline
    \end{tabular}
    \caption{Full Persona Classification Metrics}
    \label{tab:full-metrics}
\end{table}

\end{document}